\documentclass{article}  % report/book messes up chapter numbering

\usepackage{natbib}
\title{Formal Language Theory Meets Modern NLP}
\author{William Merrill\\
        \texttt{willm@allenai.org}}
\date{\today}

\usepackage[utf8]{inputenc}
\usepackage{xcolor} % for colourful comments
\usepackage{amsmath}
\usepackage{amssymb}
\usepackage{amsthm}
\usepackage{tikz}
\usepackage[hidelinks]{hyperref}
\usepackage{mathtools}

\usetikzlibrary{arrows,automata}

\theoremstyle{definition}
\newtheorem{definition}{Definition}
\theoremstyle{plain}
\newtheorem{theorem}{Theorem}

\newtheoremstyle{named}{}{}{\itshape}{}{\bfseries}{.}{.5em}{\thmnote{#3}#1}
\theoremstyle{named}
\newtheorem*{namedtheorem}{Theorem}

\usepackage{xpatch}
\makeatletter
\AtBeginDocument{\xpatchcmd{\@thm}{\thm@headpunct{.}}{\thm@headpunct{}}{}{}}
\makeatother

\newcommand{\rank}{\mathrm{rank}}

\newcommand{\computes}[1]{\rightarrow_{#1}}

\newcommand{\bigo}{\mathcal{O}}

\DeclarePairedDelimiter\abs{\lvert}{\rvert}%
\DeclarePairedDelimiter\norm{\lVert}{\rVert}%

\newcommand{\sat}{\mathrm{s}}

\newcommand{\trans}[1]{\overset{#1}{=}}

\begin{document}

\maketitle

\tableofcontents

\section{Introduction}

NLP is deeply intertwined with the formal study of language, both conceptually and historically. Arguably, this connection goes all the way back to \citeauthor{chomsky2002syntactic}'s \textit{Syntactic Structures} in 1957. It also still holds true today, with a strand of recent works building formal analysis of modern neural networks methods in terms of formal languages.\footnote{See \citet{ackerman2020survey} for a survey of recent results.} In this document, I aim to explain background about formal languages as they relate to this recent work. I will by necessity ignore large parts of the rich history of this field, instead focusing on concepts connecting to modern deep learning-based NLP. Where gaps are apparent, I will allude to interesting topics that are being glossed over in footnotes. As a general disclaimer, I encourage the reader to seek out resources on topics I ignore, written by authors who know those areas much better than me.\footnote{Naturally, this document is informed by the works I am familiar with. If you notice topics or questions that are missing, please let me know.}

The intended audience of this document is someone with a background in NLP research, and some knowledge of theory, who is interested in learning more about how the classical theory of formal languages and automata relates to modern deep learning methods in NLP.
For the sake of readability, formal proofs will be omitted in favor of proof sketches or bare statements of results. Hopefully, it will be a useful starting point that leads the reader to other resources and questions.\footnote{One classic textbook for automata theory is \citet{minsky1967computation}.}

\subsection{Basic definitions}

\paragraph{Formal language} Whereas arithmetic studies numbers and the relations over them, formal language theory is based around the study of \textit{sets of strings}. Familiar to computer scientists, a string is simply a finite sequence of tokens from some finite alphabet $\Sigma$. For example, say $\Sigma = \{a, b\}$. Then $ab$, $aabb$, and $bbaba$ are examples of strings over $\Sigma$. Each string has a finite length, but there are in fact countably infinite strings over $\Sigma$. We denote this infinite set of all strings over $\Sigma$ as $\Sigma^*$. A \emph{formal language} is a (usually infinite) set of strings, i.e. a subset of $\Sigma^*$. The set of all languages over $\Sigma$ is \emph{uncountably} infinite.\footnote{This follows from Cantor's Theorem, since the power set of $\Sigma^*$ has cardinality $2^{\aleph_0}$.}

\paragraph{Grammar} Intuitively, a ``grammar'' is the finite set of rules that describes the structure of an infinite language. In formal language theory, a \textit{grammar} can be seen as a computational system (automaton) describing a language. In particular, we will concern ourselves with three types of grammars: 
\begin{enumerate}
    % TODO: Recognize vs. decide
    \item A \textit{recognizer} $R$ is an automaton that maps from $\Sigma^* \to \{ 0, 1 \}$. $R$ implicitly defines a language $L$ where $x \in L$ iff $R(x) = 1$.
    \item A \textit{transducer} $T$ is an automaton that maps strings over one alphabet to strings over another: i.e., $\Sigma_1^* \rightarrow \Sigma_2^*$. $T$ can be seen as implementing any sequence transduction task, such as translation or mapping linguistic utterances to logical forms.
    \item An \textit{encoder} $E$ is an automaton that maps from $\Sigma^* \to \mathbb{Q}^k$, where $\mathbb{Q}^k$ is the $k$-dimensional vector space of rational numbers. An encoder can be seen as a device building a latent representation of some linguistic input.
\end{enumerate}
A \emph{grammar formalism} is a system for specifying grammars implementing these tasks. Some examples that may be familiar to an NLP audience are finite-state automata, context-free grammars, and combinatory category grammar. As you continue through this document, it may help to keep in mind two different types of questions for each formalism:
\begin{itemize}
    \item \textbf{Declarative} How are ``rules'' of a grammar specified in this formalism?
    \item \textbf{Procedural} What is the algorithm by which a grammar written in this formalism can be used to recognize, transduce, or encode a string?
\end{itemize}

With these preliminaries in place, we shall discuss a variety of formal language classes and their accompanying grammar formalisms.
\section{Regular languages} \label{sec:regular}

The most basic class of languages we will discuss is the \textit{regular languages}.\footnote{It should be noted that \textit{subregular} languages are also well studied, for example as models of natural-language phonology \citep{Heinz2011TierbasedSL}.} 
On a historical note, the genesis of theory around regular languages is intimately tied to the study of artificial neural networks. Inspired by neural circuitry, \citet{mcculloch43a} were the first to propose a model of artificial neural networks that resembled something like a finite-state automaton. Subsequently, these ideas were further developed by \citet{kleene1956automata} and \citet{minsky1956some}, forming the basis of the modern theory of regular languages that will be presented here.\footnote{I highly recommend consulting \citet{forcada2000neural} for about the connections of artificial neural networks and early automata theory.}

To a modern reader, the most natural way to define the regular languages is likely as the languages that can be recognized by regular expressions. Formally, regular expressions are an algebraic system for describing languages built around the union, concatenation, and Kleene star ($^*$) operations:\footnote{We can also write parentheses in regular expressions to disambiguate the order of operations.}

\begin{definition}[Regular languages]
For $\sigma \in \Sigma$, define $L(\sigma) = \{\sigma\}$. The regular language $L(e)$ for an expression $e$ is defined recursively:
\begin{enumerate}
    \item Union: $L(e|d) = L(e) \cup L(d)$.
    \item Concatenation: $L(ed) = \{x_ex_d \mid x_e \in L(e), x_d \in L(d) \}$.
    \item Kleene star: $L(e^*) = \bigcup_{i=0}^\infty L(e)^i$, where the Cartesian power $L(e)^i$ yields strings, e.g. $\{a,b\}^2 = \{aa, ab, ba, bb\}$.
\end{enumerate}
\end{definition}

For example, $ab^*$ is a regular expression defining the regular language of one $a$ followed by $0$ or more $b$'s, i.e., $\{a, ab, abb, \cdots \}$. Notice that, in \autoref{fig:ab*}, we draw a finite-state automaton that recognizes $ab^*$. This is an intentional choice, since it turns out that finite-state automata and regular expressions are actually two sides of the same coin. We can formally definite a deterministic finite-state automaton (DFA) as follows:

\begin{definition}[DFA] \label{def:dfa}
A DFA is a tuple $\langle \Sigma, Q, q_0, \delta, F \rangle$ where $\Sigma$ is the vocabulary, $Q$ is a finite set of states, $q_0 \in Q$ is an initial state, $\delta : \Sigma \times Q \to Q$ is the transition function, and $F \subseteq Q$ is a set of accepting states.
\end{definition}

Viewing a DFA $A$ as a language recognizer, we will now discuss the algorithm for accepting or rejecting an input string, assuming some level of familiarity.
The DFA starts in the state $q_0$. Given an input string, it reads one token $\sigma$ at a time. Letting $q$ be the current state, each token causes $A$ to \emph{transition} to a new state $q' = \delta(\sigma, q)$, which we write as $q \trans{\sigma} q'$. The machine \emph{accepts} a string $x$ of length $n$ if it ends in an \textit{accepting state} after the final token, i.e., if there exists a sequence of states $\{q_i\}_{i=1}^n$ such that
\begin{equation}
q_0 \trans{x_1} q_1 \trans{x_2} \cdots \trans{x_n} q_n \in F .
\end{equation}
\noindent The set of languages recognized by $A$ is the set of strings that it accepts. We will refer to the number of states $\abs{Q}$ as the \emph{size} of $A$.

An important classical result is that the regular languages are equivalent to the languages that can be accepted by finite-state automata:

\begin{figure}
    \centering
    \begin{center}
    \begin{tikzpicture}[->,>=stealth',shorten >=1pt,auto,node distance=2.8cm, semithick]
    \node[initial,state]        (0) {$q_0$};
    \node[accepting,state]          (1)[right of=0] {$q_1$};
    \path (0) edge node {$a$} (1)
          (1) edge [loop right] node {$b$} (1);
    \end{tikzpicture}
    \end{center}
    \caption{DFA that recognizes the regular language $ab^*$. Accepting states ($\in F$) are indicated with a double circle border. To simplify the graph, we assume undefined transitions terminate the computation, returning $0$.}
    \label{fig:ab*}
\end{figure}
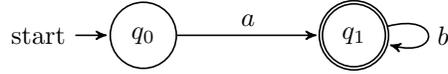

\begin{namedtheorem}[Kleene's ] \label{thm:kleene}
For all languages $L$, $L$ is recognizable by a DFA iff $L$ is recognizable by a regular expression.\footnote{In fact, this applies for both deterministic and nondeterministic finite-state acceptors, as these two classes are equivalent \citep{Rabin1959FiniteAA}. Nondeterministic automata, while not as computationally efficent, can be used to concisely describe regular languages.}
\end{namedtheorem}

An important generalization of the DFA is the nondeterministic finite automaton (NFA). At a high level, NFAs relax the transition function $\delta$ to be a relation, meaning that more than one transition can be possible for any token. A string is accepted if \emph{any} sequence of transitions leads to an accepting state.\footnote{We will discuss nondeterministic automata in greater detail in \autoref{sec:data-structures}.} Extending Kleene's Theorem, the set of languages recognizable by NFAs is also the regular languages. Thus, in the finite-state setting, nondeterminism does not introduce any additional expressive power. However, the size of a ``determinized'' DFA generated from an arbitrary NFA can be exponentially larger than the original NFA.

In summary, regular expressions, DFAs, and NFAs all converge to the same recognition capacity: the regular languages.
There is a lot more to be said about the theory and practice of regular languages. For a nice overview of the theory from first principles, including the Myhill-Nerode Theorem and language derivatives, see \citet{wingbrant2019regular}.

\subsection{Learning regular languages}

As our focus here is on the relevance of formal language theory for machine learning, we will discuss foundational work on learning regular languages from data. The most basic question we might want to know about this is how difficult it is to learn a DFA recognizer from positive and negative data. By this, we mean a finite sample of strings in $L$ as well as strings in $\Sigma^* \setminus L$. It is always possible to construct a DFA to fit such a sample: we can build a trie (tree of prefix states) mapping each full string to its label. Since such a trie is a special case of a DFA, we have a construction for a DFA recognizer whose size is linear in the size of the sample. However, this large DFA will not generalize to new samples. Inspired by Occam's razor, we instead might hope to find the \emph{smallest} DFA matching our data set, hoping that this lower complexity automaton would generalize well to other strings in $L$. This turns out to be a computationally intractable problem: \citet{Gold_1978} proved that, in fact, it is NP-hard. 

However, all is not lost. Considering a slightly different formal setup, \citet{Angluin1987LearningRS} introduced the $L^*$ algorithm, which finds minimal DFAs for regular languages from a polynomial number of \emph{membership} and \emph{equivalence} queries. Let $L$ be a ``ground truth'' regular language. A membership query takes as input a string, and returns whether it is in $L$. An equivalence query takes as input a hypothesis $L'$, represented by a finite-state machine. If $L = L'$, the query returns $\emptyset$; otherwise, it returns a ``counterexample'': a string accepted by one machine, but not the other. The $L^*$ algorithm uses these queries to converge to a minimal finite-state machine for $L$ in polynomial time.
\section{Data structures} \label{sec:data-structures}

Finite-state automata represent the limits of computation with only finite memory available. With additional memory, a recognizer is capable of specifying more complex languages.
One way to view more complicated formal grammars, then, is as finite-state automata augmented with various kinds of data structures, which introduce different kinds of unbounded memory. This is the view we will take for introducing deterministic pushdown and counter automata, two useful abstractions for understanding the internal computation of RNN architectures.

\begin{definition}[Data structure] \label{def:datastruct}
A data structure $\mathcal D$ is a tuple $\langle c_0, U, r \rangle$.
\begin{enumerate}
    \item The \emph{null} data structure $c_0$ represents the base case for a newly constructed object. For example, this could be an empty stack.
    \item The \emph{update operations} $U$ are a finite set of computable functions $u : C \rightarrow C$, where $C$ is the space of data structure states. We can interpret each $u$ as taking an old state $c$, and returning an updated one $c'$. For example, for a binary stack, $U$ could be $\{ \textsc{noop}, \textsc{pop}, \textsc{push}_0, \textsc{push}_1 \}$. The update $\textsc{pop}$ would remove the top element, e.g. $1001 \mapsto 001$.
    \item The \emph{readout operation} $r : C \rightarrow R$ is a computable function that produces finite summaries $r(c)$ from data structure states $c$. Importantly, the range $R$ must be finite. For a stack, we can take $r$ to return the top $k$ elements of the stack, for some finite $k$.
\end{enumerate}
\end{definition}

This definition of a data structure provides a generic way to formalize unbounded memory that can interface with a finite-state machine. We now define an automaton that ``controls'' the data structure, i.e. updates it and reads from it according to the input string. Notice how the following generalizes the finite-state recognizer from \autoref{def:dfa}:

\begin{definition}[$\mathcal D$-automaton]
Let $\mathcal D$ be a data structure $\langle c_0, U, r \rangle$ as in \autoref{def:datastruct}.
A $\mathcal D$-automaton is a tuple $\langle \Sigma, Q, q_0, \upsilon, \delta, F \rangle$ with
\begin{enumerate}
    \item A finite alphabet $\Sigma$
    \item A finite set of states $Q$
    % \footnote{The original definition \cite{fischer1968counter} distinguishes between ``autonomous" and ``polling" states, a distinction that is vacuous for real-time machines.}
    \item An initial state $q_0 \in Q$
    \item A memory update function $\upsilon : \Sigma \times R \times Q \rightarrow U$
    \item A state transition function $\delta : \Sigma \times R \times Q \rightarrow Q$
    \item An acceptance mask $F \subseteq R \times Q$
\end{enumerate}
\end{definition}

What does it mean to ``compute'' with a $\mathcal D$-automaton? The definition of language recognition is quite similar to that of a DFA, except that the steps can update and read from the data structure. Let's make this precise.
A full configuration for a $\mathcal D$-automaton is a data structure configuration $c$ as well as a finite state $q$, Given $\langle c, q \rangle \in C \times Q$ and an input symbol $\sigma \in \Sigma^*$, we define the transition of the finite state $q$ and data structure configuration $c$:
\begin{align} 
    c' &\trans{\sigma} \upsilon(\sigma, r(c), q) (c) \label{eq:double-fn} \\
    q' &\trans{\sigma} \delta(\sigma, r(c), q) .
\end{align}
\noindent Note that $\upsilon(\sigma, r(c), q)$ returns a function in \eqref{eq:double-fn}, which explains the iterated function application. We will overload $\trans{\sigma}$ to full machine configurations, i.e., $\langle c, q \rangle \trans{\sigma} \langle c', q' \rangle$ as defined above.
Recall that $c_o$ is the null data structure, and $q_0$ is the initial state.
We say that the machine accepts a string $x \in \Sigma^*$ of length $n$ if there exists a sequence of configurations $\{\langle c_i, q_i \rangle \}_{i=1}^n$ such that
\begin{equation} \label{eq:acceptance}
    \langle c_0, q_0 \rangle 
    \trans{x_1} \langle c_1, q_1 \rangle
    \trans{x_2} \cdots
    \trans{x_n} \langle c_{n}, q_{n} \rangle ,
\end{equation}
\noindent and $\langle r(c_n), q_n \rangle \in F$. Informally, this says that the machine accepts $x$ if its deterministic computation sequence ends in an accepting configuration.

\subsection{Stacks and counters}

This general framework allows us to formalize recognizers using two types of data structures relevant to recent theoretical work on RNNs: stack and counter automata. At a high level, we get these machines by letting $\mathcal D$ be either a stack or a vector of integer-valued counter variables.

\begin{definition}[Stack automaton]
Let $C = \{0, 1\}^*$ and $c_0 = \epsilon$ (the empty string).
For a string $c$, let $c_{2:}$ denote the string with the first token deleted.
We define the set $U = \{ \textsc{noop}, \textsc{pop}, \textsc{push}_0, \textsc{push}_1 \}$ where
\begin{align}
    \textsc{noop}(c) &= c \\
    \textsc{pop}(c) &= c_{2:} \\
    \textsc{push}_0(c) &= 0 c \\
    \textsc{push}_1(c) &= 1 c .
\end{align}
\noindent We define $r$ to return the $k$ topmost elements of the stack: $r(c) = c_{:k}$.
\end{definition}

Another name for a stack automaton is a \emph{deterministic pushdown automaton}. Intuitively, this machine works by pushing and popping different symbols onto a stack. The topmost contents of the stack can in turn inform the behavior of the automaton.

\begin{definition}[Counter automaton]
Let $C = \mathbb{R}^k$, and $c_0$ be the $k$-dimensional zero vector. We set $U = \{ {-}1, {+0}, {+}1, {\times}0 \}^k$, where each element is a function, e.g. ${+}1(x) = x + 1$. We let $r(c)$ be the broadcasted ``zero-check'' function, i.e., for $1 \leq i \leq n$:
\begin{equation}
    r(c)_i = 
    \begin{cases}
        0 & \textrm{if} \; c_i = 0 \\
        1 & \textrm{otherwise.}
    \end{cases}
\end{equation}
\end{definition}

This definition is equivalent to the real-time, deterministic counter automata referred to in other works \citep{merrill2020linguistic, fischer1968counter}. Intuitively, these automata resemble stack automata, except that the stack data structure has been replaced by a finite number of ``integer'' variables. At each computational step, these variables can be added to or subtracted from. The automaton can condition its behavior on whether or not the counters are $0$.\footnote{\citet{merrill2020linguistic} shows various generalized operations over counters, such as adding or thresholding at arbitary integers, that can be introduced without increasing the expressive power of the automaton.}

\subsection{Power of unbounded memory}

The unbounded memory introduced by counters or a stack allows these automata to handle more languages than finite-state recognizers. For example, consider the language $a^nb^n$, representing the set of strings with a certain number of $a$'s followed by a matching number of $b$'s. A finite-state recognizer cannot accept $a^nb^n$ because doing so requires ``remembering'' the number of $a$'s, which is unbounded as $n$ gets arbitrarily large.

With a counter automaton, however, we can simply add $1$ to a counter when we see $a$, and subtract $1$ when we see $b$. At the end of the string, we accept if the counter is back at $0$.\footnote{Using finite state, we also track whether we have seen $ba$ at any point. This lets us constrain acceptance to strings of the form $a^*b^*$.} Thus, a counter-based strategy provides one way to handle this language.

With a stack automaton, we have a different way to create a recognizer for $a^nb^n$. We push each $a$ onto the stack. For a $b$, we pop if the stack contains $a$, and raise an error if it contains a $b$ or is empty. Finally, at the end of the string, we accept if the stack is empty.

Thus, both stack and counter memory provide ways to accept the language $a^nb^n$. We will see, though, that other languages differentiate the abilities of these modes of computation.

\subsection{Power of counters}

Parentheses languages pose a problem for counter machines, whereas they are easy to accept for a stack automaton. We define the Dyck (parentheses) languages by the following context-free grammar:

\begin{definition}[$k$-Dyck language]
For any $k$, define the $k$-Dyck language $D_k$ as the language generated by the following context-free grammar:
\begin{align}
    \mathrm{S} &\rightarrow \mathrm{S} \mathrm{S} \; | \; \epsilon \\
    \mathrm{S} &\rightarrow (_i \mathrm{S} )_i \quad (\textrm{for all} \; 1 \leq i \leq k) ,
\end{align}
where $\epsilon$ is the empty string. Note that $(_i$ and $)_i$ refer to tokens in $\Sigma$.
\end{definition}

Intuitively, $D_k$ represents the language of well-balanced parentheses, with $k$ types of parenthetical brackets. Both counters and a stack can be used to accept $D_1$, as $D_1$ is quite similar to $a^nb^n$. However, for $k \geq 2$, counter automata cannot recognize the language. The root of the problem is that a recognizer must model not just the number of open parentheses at any point, but also remember the full sequence of parentheses. A stack is naturally suited to do this---just push each open bracket---but a counter machine will quickly run out of memory. To be more precise, the number of memory configurations of a counter automaton is bounded polynomially in the input length, whereas $k$-Dyck for $k \geq 2$ requires \textit{exponentially} many memory configurations.

% TODO: move this to CFG section?

An interesting aside here is that, in some sense, Dyck structure is representative of all hierarchical structure. This is formalized by the following theorem:

\begin{theorem}[\citealt{Chomsky1963TheAT}] \label{thm:schutzenberger}
Let $L$ be any context-free language.
Then $L$ is the homomorphic image of a Dyck language intersected with a regular language, i.e.
for some $k$, regular language $R$, and homomorphism $h$,
\begin{equation*}
    L = h(D_k \cap R) .
\end{equation*}
\end{theorem}

Thus, not being able to accept Dyck languages is truly a fundamental weakness of counter automata as models of hierarchical structure, which is widely believed to be an important feature of natural language. We will say more about the relationship between context-free languages and natural language in \autoref{sec:cfgs++}.

% https://www.math.uni-heidelberg.de/logic/ss17/formsprach/fs10-schuetzenberger-41.pdf

\subsection{Power of a stack}

The set of languages recognizable by a stack automaton is called the \emph{deterministic context-free languages} (DCFLs), and forms a strict subset of the context-free languages (CFLs) proper.\footnote{An example of a CFL that is not deterministic is $\{a^nb^n \mid n \geq 0 \} \cup \{a^nb^{2n} \mid n \geq 0\}$. See \citet{power2002} for a nice proof of this.} Compared to CFLs, DCFLs always permit an \emph{unambiguous} grammar: meaning a parsing system where each string is assigned exactly one tree.\footnote{The converse is not always true: some unambiguous CFLs are not deterministic. Intuitively, this is related to the concept of local ambiguity, where a full string can have an unambiguous parse, but ambiguity exists for building intermediate parses when reading the string left-to-right. In natural language, these kind of ambiguities can lead to \href{http://www-personal.umich.edu/~jlawler/gardenpath.pdf}{garden-path sentences} like \emph{The old man the boat}.}
On the other hand, the Dyck languages are DCFLs, and we have already seen that, in some sense, they capture the fundamental structure of all CFLs according to \autoref{thm:schutzenberger}.

Despite this power, stack automata suffer from their own weaknesses relative to counter automata. There is in fact no strict containment relation between counter languages and DCFLs, or CFLs for that matter. An illustrative example is the language $a^nb^nc^n$, which can be accepted by a machine with two counters. The first one counts up for $a$'s and down for $b$'s, and the second one counts up for $b$'s and down for $c$'s. A stack automaton, however, cannot do this. The intuition is that each $a$ can only be closed by one item, so once we have matched it to some $b$, we cannot match it to any $c$.\footnote{This can be formalized by using the pumping lemma to prove that $a^nb^nc^n$ is not a context-free language (and thus not a deterministic context-free language).}

\subsection{Nondeterminism}

In \autoref{sec:regular}, we remarked that NFAs are equivalent in capacity to DFAs. Interestingly, this correspondence between nondeterministic and determistic automata does not hold for memory augmented automata. Informally, nondeterminism means that that transitions allow many possibilities instead of just one. A string is accepted if \emph{any} choice of valid transitions ends in an accepting state. Formally, we relax the transition and state update functions to \emph{relations} $\upsilon \subseteq \Sigma \times R \times Q \times U$ and $\delta \subseteq \Sigma \times R \times Q \times Q$.

From a procedural point of view, we generalize the automaton transition rule as follows.
We say that $\langle c, q \rangle \trans{\sigma} \langle c', q' \rangle$ if there exists $u \in U$ such that:
\begin{align}
    c' &= u(c) \\
    \langle \sigma, r(c), q, u \rangle &\in \upsilon \\
    \langle \sigma, r(c), q, q' \rangle &\in \delta .
\end{align}
\noindent We say that a nondeterministic automaton accepts a string if there exists some sequence of valid transitions that reaches an accepting state. The language recognition capacity of the machine is defined as the set of strings that are accepted according to this overloaded definition.

This relaxation means that our automaton no longer must proceed in a linear sequence of transitions, but instead can ``backtrack'' across many possible transition sequences. This essentially allows the automaton to traverse an exponential space of possible computational paths while deciding to accept a string, increasing expressive power while making the machine less practical to implement. In the case of stack automata, the recognition capacity is expanded from the DCFLs to the CFLs proper.\footnote{At the time of writing, I am not immediately sure what the capacity of the nondeterministic counter automaton is. Let me know if you have thoughts about it!}

\subsection{Learning memory-augmented automata}

One interesting question is how methods for learning finite-state automata can be extended to more complex automata with counter or stack memory. The field of grammar induction consists of a family of problems related to this idea, attempting to infer context-free grammars or other models of grammar from data using algorithms similar to $L^*$. See \citet{grammaticalInference} for a textbook providing a thorough overview of this extensive field.

More recently, a thread of work in deep learning has attempted to re-frame data structures like stacks as differentiable objects that can be seamlessly integrated with standard gradient-based optimization \citep{graves2014neural, grefenstette2015learning, suzgun2019memoryaugmented}. The data structure can be directly controlled by an RNN in a way that resembles a $\mathcal D$-automaton. Since the data structure is differentiable, gradients can be backpropagated through it to update the network's parameters. This approach combines the structural biases and interpretability of data structures with the practical utility of high-dimensional optimization for learning. When done well, it produces models that learn how to use data structures to solve a particular problem, perhaps similarly to a human programmer. However, there are challenges in defining the differentiable data structures in ways that are amenable to learning and efficient at train time \citep{Hao_2018}.
Empirically, stack-augmented networks have been shown to perform well on synthetic tasks requiring strong hierarchical representations \citep{grefenstette2015learning, suzgun2019memoryaugmented} as well as on some natural language tasks \citep{yogatama2018memory, merrill-etal-2019-finding}.
Theoretically, \citet{stogin2020provably} proved that stack-augmented RNNs are stable, meaning that the stack state remains close to that of some stack automaton.
Although most work incorporating stack memory into RNNs has used deterministic controllers, \citet{dusell2020learning} recently developed an algorithm for training a neural network simulating a nondeterministic stack automaton. In principle, this approach can also be applied to other data structures beyond stacks, such as Turing machines \citep{graves2014neural}. In \autoref{sec:saturation}, we will develop the perspective that the LSTM gating mechanism can be viewed similarly to a differentiable approximation of a counter data structure.
\section{Chomsky hierarchy} \label{sec:cfgs++}

\begin{figure}
    \centering
    \includegraphics[width=.6\textwidth]{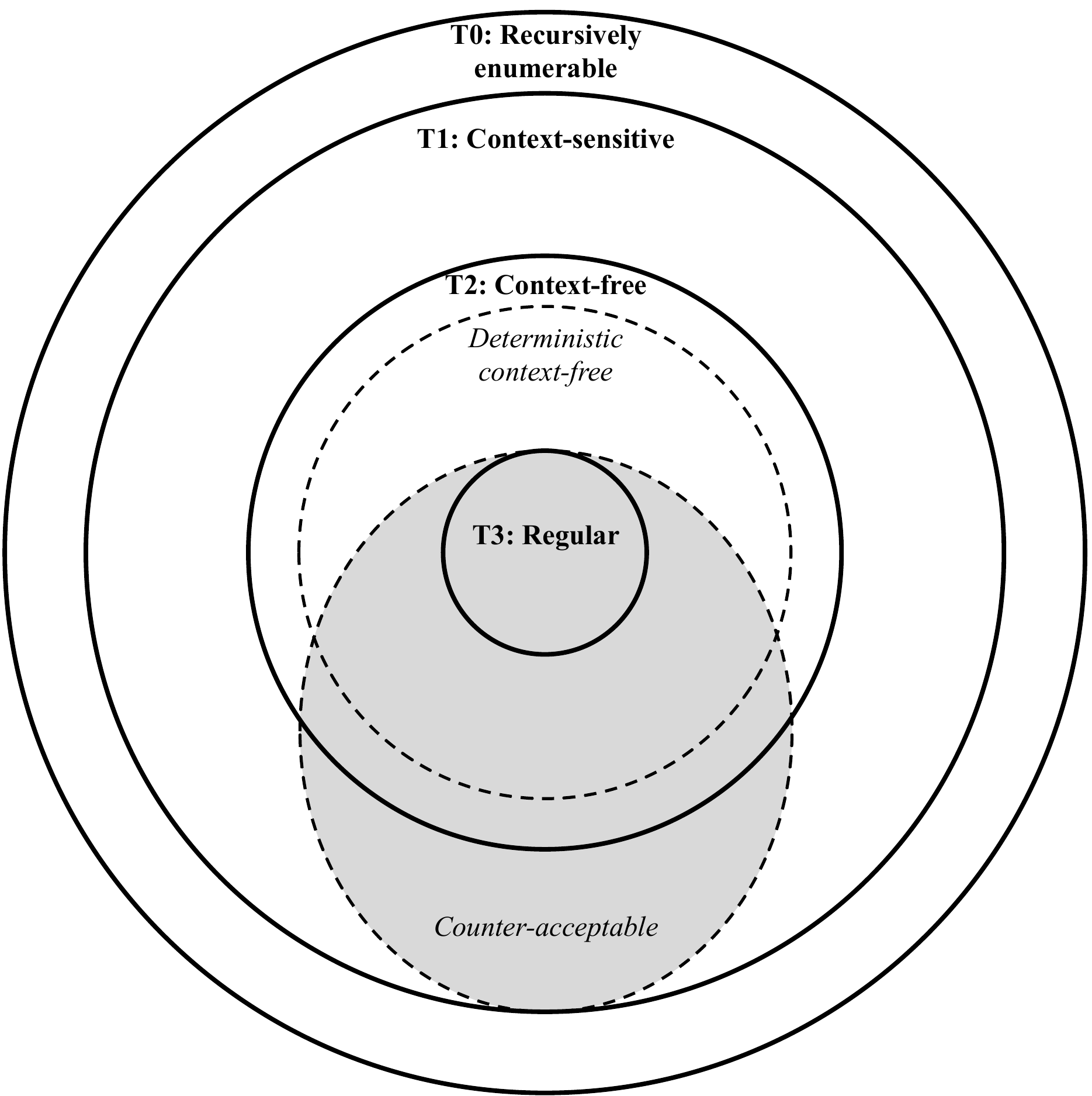}
    \caption{The Chomsky hierarchy superimposed with some additional classes of languages we have discussed. Recall that the capacity of a stack automaton is the deterministic context-free languages, and the capacity of a counter automaton is the counter(-acceptable) languages.}
    \label{fig:hierarchy}
\end{figure}

The reader may be familiar with the Chomsky hierarchy \citep{Chomsky1956Three, Chomsky1959OnCF}: a classical framework for arranging the syntactic complexity of formal languages.\footnote{A good reference on the Chomsky hierarchy is \citet{hopcroftBook}.} The Chomsky hierarchy arranges classes of formal languages in concentric sets based on the complexity of the grammars needed to parse them. As shown in \autoref{fig:hierarchy}, the results we have so far explored all lie in the lower levels of the classes hierarchy.
The regular languages from \autoref{sec:regular} form Type 3: the least complex level of the classical Chomsky hierarchy. The memory-augmented automata discussed in \autoref{sec:data-structures}, denoted in \autoref{fig:hierarchy} with dashed lines, fall withing Type 2 and Type 1. The Dyck languages fall within the deterministic context free languages, intersecting the counter-acceptable ones ($D_1$), but not fully contained ($D_k$ for $k > 1$).

What goes on in the higher levels of the hierarchy? Type 2 consists of the context-free languages, which are acceptable by a nondeterministic stack automaton. In this context, nondeterministic means that the the transitions of the machine are not functions, but relations. The machine accepts a string if any choice of transitions over the string ends in an accepting configuration. As discussed in \autoref{thm:schutzenberger}, context-free languages can be thought of as languages with nested hierarchical structure. Type 1 consists of the context-sensitive languages, corresponding to the languages acceptable by a nondeterministic Turing machine with $O(n)$ space. Finally, at the top of the hierarchy, we have Type 0, which corresponds to the recursively enumerable functions. The acceptors for this class of languages are arbitrary Turing machines, establishing them at the top of the hierarchy.

A big question in mathematical linguistics is the place of natural language within this hierarchy. Context-free formalisms describe much of the hierarchical structure of natural languages. However, the presence of various types of cross serial dependencies (which do not easily fall into projective dependency trees) lead many linguistics to believe that natural language is \emph{mildly context-sensitive} \citep{shieber1985evidence}.\footnote{The canonical example of this based purely on syntax is Swiss German, where center-embedded noun verb phrases embed serially rather than nesting. However, a mildly context-sensitive analysis also makes recovering deep structural (i.e. semantic) dependencies easier in \emph{any} language.} In terms of the Chomsky hierarchy, this means that it falls in a class between the context-free and context-sensitive classes, but close enough to the context-free class such that parsing remains tractable. There is much debate about what the right formalism is: two contenders are Tree Adjoining Grammar \citep[TAG; ][]{joshi1975tree} and Combinatory Category Grammar \citep[CCG; ][]{steedman1996very}.

Another interesting extension of the Chomsky hierarchy is hypercomputation \citep{siegelmann1995computation}. Hypercomputation refers to models of computation that are more powerful than Turing machines. While this may seem to violate the Church-Turing thesis that Turing machines represent a universal model of effective computation, it does not, because all known models of hypercomputation are \emph{infinite} machines. The standard theory of computability, on the other hand, is concerned with the ability of \emph{finite} machines to model infinite sets of strings. Various models of hypercomputation exist, and can model languages outside of Type 0. In fact, Type 0 only contains countably infinite languages (since there are countably infinite Turing machines), whereas the set of all formal languages is uncountably infinite. Thus, in some sense, ``most'' formal languages are not acceptable by any computable grammar. We will return to the notion of hypercomputation in \autoref{sec:hypercomputation}, where we explore interesting theoretical connections to the capacity of real-valued neural networks.
\section{Weighted finite-state automata} \label{sec:wfas}

So far, we have mostly discussed different types of recognizers for formal languages. We now turn to discussing a type of encoder, a weighted finite-state automaton (WFA), which generalizes a finite-state recognizer. The WFA uses a finite-state machine to encode a string into a number, rather than producing a boolean decision for language membership.

Let $\langle \mathbb{K}, \oplus, \otimes \rangle$ be a semiring.\footnote{For concreteness, you can imagine the rational field $\langle \mathbb{Q}, +, \cdot \rangle$, but WFAs are often defined over other semirings, such as $\langle \mathbb{Q}, \max, + \rangle$.}
We use this semiring to define a WFA as a finite-state machine augmented with weighting functions for each transition. To score a string, we consider each path that the string induces through the automaton. We score a path by ``multiplying'' ($\otimes$) the transitions along it. We then score the full string by ``summing'' ($\oplus$) the scores for all possible paths. The notion of initial and accepting states are generalized to allow special weights that can be multiplied for starting and ending in each state. Let's go through this a little more formally.

Let $\langle \mathbb{K}, \oplus, \otimes \rangle$ be some semiring. A WFA $W$ consists of an alphabet $\Sigma$, set of states $Q$, and weighting functions $\lambda, \tau, \rho$ defined as follows:

\begin{enumerate}
    \item Initial state weights $\lambda : Q \rightarrow \mathbb{K}$
    \item Transition weights $\tau : Q \times \Sigma \times Q \rightarrow \mathbb{K}$
    \item Final state weights $\rho : Q \rightarrow \mathbb{K}$    
\end{enumerate}

\noindent Let $q \computes{\sigma} q'$ denote a transition from $q$ to $q'$ licensed by token $\sigma$. The weighting functions of a WFA are used to encode any string $x \in \Sigma^*$ as follows:

\begin{definition}[Path score]
Let $\pi$ be a path of the form $q_0 \computes{x_1} q_1 \computes{x_2} \cdots \computes{x_t} q_t$ through WFA $A$. The score of $\pi$ is given by
\begin{equation*}
    W(\pi) = \lambda(q_0) \otimes \left( \bigotimes_{i=1}^t \tau(q_{i-1}, x_i, q_i) \right) \otimes \rho(q_t) .
\end{equation*}
\end{definition}
\noindent By $\Pi(x)$, denote the set of paths producing $x$.

\begin{definition}[String encoding] \label{def:string-score}
The encoding computed by a WFA $A$ on string $x$ is 
\begin{equation*}
    W(x) = \bigoplus_{\pi \in \Pi(x)} A(\pi) .
\end{equation*}
\end{definition}

If we restrict the $\mathbb{K}$ to $\langle \{0, 1\}, \vee, \wedge \rangle$, then we get nondeterministic finite recognizers, whose recognition capacity is the regular languages. If we use the rational field as our semiring, the WFAs compute a class of function called the \textit{rational series}. In this sense, the rational series are the class of encoding functions that generalize the regular languages.

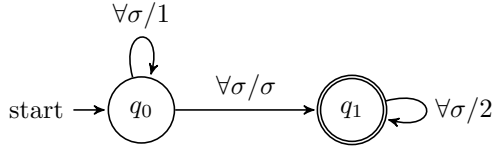
\begin{figure}
    \centering
    \begin{tikzpicture}[->,>=stealth',shorten >=1pt,auto,node distance=2.8cm, semithick]
    
    \node[initial,state]           (0) {$q_0$};
    \node[accepting,state]         (1) [right of=0] {$q_1$};
    
    \path (0) edge [loop above] node {$\forall \sigma/1$} (0)
              edge              node {$\forall \sigma/\sigma$} (1)
          (1) edge [loop right] node {$\forall \sigma/2$} (1);
    
    \end{tikzpicture}
    \caption{A WFA over$\langle \mathbb{Q}, +, \cdot \rangle$ mapping binary strings to their numeric value. This can be extended for any base $>2$.
    Let $\forall \sigma/w(\sigma)$ denote a set of transitions consuming each token $\sigma$ with weight $w(\sigma)$. We use standard DFA notation to show initial weights $\lambda(q_0) = 1, \lambda (q_1) = 0$ and accepting weights $\rho(q_0) = 0, \rho(q_1) = 1$.
    % \citet{CFG-WFA} present a similar construction.
    }
    \label{fig:binary}
\end{figure}

The rational series are counterintuitively powerful, capable of producing encodings that are ``stateful'', i.e. depend on more than finite context. For example, consider the function mapping a string $\{0, 1\}$ to the value that it represents in binary. For example, $101 \mapsto 5$. This function takes exponentially many output values, in the input string length. Nevertheless, \autoref{fig:binary} shows a WFA that computes it.

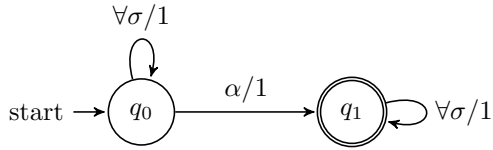
\begin{figure}
    \centering
    \begin{tikzpicture}[->,>=stealth',shorten >=1pt,auto,node distance=2.8cm, semithick]
    
    \node[initial,state]           (0) {$q_0$};
    \node[accepting,state]         (1) [right of=0] {$q_1$};
    
    \path (0) edge [loop above] node {$\forall \sigma/1$} (0)
              edge              node {$\alpha/ 1$} (1)
          (1) edge [loop right] node {$\forall \sigma /1$} (1);
    
    \end{tikzpicture}
    \caption{A WFA over $\langle \mathbb{Q}, +, \cdot \rangle$ that counts occurrences of the unigram $\alpha$. This can be generalized with $n+1$ states to count all occurences of an $n$-gram. Each occurence of $\alpha$ produces a path with score $1$; thus, the string score is the number of occurences of $\alpha$.}
    \label{fig:sigma-CM}
\end{figure}

Another useful WFA to keep in mind is the $n$-counter. This machine uses $n+1$ states to count all the occurrences of a specific $n$-gram. A $1$-counter is shown in \autoref{fig:sigma-CM}. Thus, there is a natural connection between WFAs and counter machines: WFAs can simulate counting where the update to the counters cannot be conditioned by their current values. As discussed in \citet{merrill-etal-2020-formal}, this relationship is useful for developing a typology of RNN architectures. \citet{merrill-etal-2020-formal} also explore the power of a WFA encoder as a recognizer where various types of decoding functions are used to accept or reject based on the latent encoding.

\subsection{Hankel matrices}

The Hankel matrix is an infinite matrix that can be used to represent any formal language. It has many appealing theoretical properties that relate directly to WFAs.

\begin{definition}
Given a function $f : \Sigma^* \rightarrow \mathbb{K}$ and two enumerations $\alpha, \omega$ of $\Sigma^*$, the Hankel matrix $H_f$ is defined at each coordinate $i,j \in \mathbb{N}$ as:
\begin{equation*}
    [H_f]_{ij} = f(\alpha_i \omega_j) .
\end{equation*}
\end{definition}

This matrix has an elegant theoretical connection to the theory of WFAs:

\begin{theorem}
Consider $f : \Sigma^* \rightarrow \mathbb{Q}$. Then $H_f$ has finite rank iff there exists a WFA over the rational field computing $f$. Further, $\mathrm{rank}(H_f)$ is equal to the number of states of this WFA.
\end{theorem}

This is a perhaps surprising connection between rank, a notion in linear algebra, and the discrete states of a finite-state machine.
It justifies using a finite state-machine to describe an infinite matrix, or vice versa. If $A$ is a sub-block of $\rank(H_f)$, $\rank(A) \leq \rank(H_f)$. Thus, an empirical Hankel sub-block $\hat H_f$ suffices to get a lower bound on the size of the finite-state machine computing $f$.

\subsection{Learning WFAs}

The correspondence between Hankel matrices and WFAs also suggests a natural \textit{spectral algorithm} for learning WFAs from data \citep{Hsu_2012}. The algorithm involves building a large empirical sublock $\hat H_f$ of the Hankel matrix $H_f$. It then uses a singular value decomposition (SVD) to solve for the parameters of a WFA in terms of $\hat H_f$. If $\hat H_f$ is large enough, i.e. has the same number of states as the underlying $H_f$, then this approach will reconstruct the underlying WFA. Variants of the algorithm exist, differing in their practical performance with low or noisy data. Since $\hat H_f$ can be built from a ``black box'' function $f$, this approach can be applied to derive WFAs approximating arbitrary behavior, such as the representations or outputs from a neural network \citep{rabusseau2018connecting}. A related theoretical work \citep{marzouk2020distance} considers the problem of deciding equivalence between a WFA and an RNN, showing it to be undecidable in general and EXP-hard over a finite support of data.

\section{Foundational neural networks results}

Having reviewed core concepts about formal languages and automata, we shift focus towards theory that centers neural networks as an object of study. We first discuss foundational results about neural networks as language recognizers. We assume some level of familiarity with modern NLP architectures like RNNs and transformers.

It is perhaps useful to have some historical perspective on the development of this area.
The earliest expressiveness results about neural networks predate the ``deep learning revolution'' of the early 2010s. In these days, neural networks were viewed in some research as an abstract model of computation rather than a practical method for learning behavior from data. Neural networks were an interesting alternative model of computation because of their distributed nature and potential resemblance to the human brain.

\subsection{Hypercomputation} \label{sec:hypercomputation}

One distinctive feature of neural networks is that their state can be viewed as real-valued, rather than discrete like in a Turing machine. \citet[\textit{inter alia}]{siegelmann1995computation, Siegelmann2004NeuralAS} explore the implications of having real-valued state in a computational system, showing how it leads real-valued RNNs to be capable of hypercomputation. In other words, they can compute functions over $\Sigma^* \rightarrow \{0, 1\}$ that a Turing machine can never compute in finite time, such as the halting problem.\footnote{Specifically, their capacity is the complexity class $\mathrm{P} / \mathrm{poly}$.}
Intuitively, the reason why this is possible is that real numbers take infinite bits to specify, and thus a description of a real-valued neural network is actually an infinitely sized object, whereas conventional computational systems are finite. In computational complexity theory, a similar intuition explains why infinite circuit families are capable of hypercomputation \citep[Chapter 6]{arora2009}. One interpretation of Siegelmann's results is that ``brainlike'' systems whose state can be specified with infinite precision may be qualitatively different than computational devices like Turing machines.

\subsection{Turing completeness} \label{turing-completeness}

By restricting the network weights to \textit{rational} numbers instead of real numbers, we make it a well-defined computational system by normal standards of computability. As shown by \citet{Siegelmann1992OnTC}, these rational RNNs are in fact Turing-complete, assuming arbitrary (but finite) precision of the rational weights, and unbounded computation time, i.e. recurrent steps.\footnote{\citet{Siegelmann1992OnTC} talk about ``linear time'', by which they mean the runtime of the neural network is linear in the runtime of the simulated Turing machine. However, the absolute number of recurrent steps is still unbounded, which is more powerful than the real-time (one step per input) use of neural networks in deep learning.} The trick behind the construction is that, since a single neuron has arbitrarily precision, we can use a small, finite number of neurons to remember a full Turing machine tape, no matter how big it gets. \citet{Siegelmann1992OnTC} define a network graph that simulates running steps of a two-stack Turing machine. Recursively running this network until the simulated machine halts allows us to compute any Turing-computable function.

\paragraph{Caveats of this result} This setting and construction differ substantially from how we think about neural networks in modern deep learning. Generally, the computation graph is unrolled proportionally to the input length, restricting us to \textit{real-time} computation.
Furthermore, the precision of the network state is bounded in practice, and attempting to encode large objects like Turing machine tapes within individual neurons will quickly run into hardware-imposed limits.
Finally, while they can be hand-constructed, it seems unlikely that constructions exploiting arbitrary precision are easily learnable by gradient-based optimization. The encoding will be highly sensitive to slight perturbations in the network parameters,\footnote{In \autoref{sec:saturation}, we will see that constructions to simulate Turing machines can actually only exist at small $\ell_2$ norm within an RNN's parameter space.} meaning that the solution corresponds to steep valleys in the parameter space. Intuitively, gradient descent in a high dimensional space is unlikely to find steep minima, an idea that various works in deep learning theory \citep[e.g.][]{keskar2017large-batch, he2019asymmetric} have explored.

% and it is unlikely that the network will learn representations requiring strong precision, corresponding to very steep minima in the loss landscape. Thus, more recent work has theorized ways of formalizing capacity that are potentially more descriptive for practical networks learned from data. Still, in my experience, intro NLP classes tend to refer vaguely to the Turing completeness of RNNs. I believe it is important to understand the sense in which RNNs are Turing-complete, since there are important differences between the formal model and the classes of networks used by modern deep learning.

\paragraph{Hierarchy} Another interesting thing to note is that there is an infinite hierarchy of complexity classes between the rational-valued networks and real-valued networks, differentiated by how many real-valued neurons are granted \citep{superHierarchy}. For example, a network with $1$ real-valued neuron is more powerful than a network with $0$, and a network with $2$ is even more powerful than the network with $1$. The increase in power can be described in terms of a generalization of Kolmogorov complexity.

\subsection{Binarized neurons} \label{sec:binarized}

Having discussed real and rational-valued neural networks, we now move to the weakest variant: networks whose weights are binarized. We can formalize this by assuming the network weights are rational or integer-valued, but that the activation function of each neuron thresholds the activation value into $\{0, 1\}$. This model of computation is much weaker than Turing-complete. In fact, binarized RNNs are equivalent to finite-state machines, which we have seen can only compute regular languages.
More recently, \citet{sima2020} introduced the analog neuron hierarchy, which explores the hierarchy between having a fully binarized RNN and a rational-valued RNN based on the number of rational neurons. Adding more neurons increases capacity, until, at $3$ rational neurons, the hierarchy collapses to the full Turing-complete capacity of rational-valued RNNs. A recent extension \citep{sima2021stronger} proves further separations between intermediate classes in this hierarchy.
This strand of work relates to saturated RNNs, which we will explore in the next section.

\section{Saturated networks} \label{sec:saturation}

As mentioned in \autoref{turing-completeness}, RNNs with unbounded time and infinitely precise states are capable of impractical behavior like simulating a whole Turing machine tape in a small number of neurons. In this section, we discuss the framework of \textit{saturated} networks.\footnote{In \citet{merrill-2019-sequential}, this was called the \emph{asymptotic network}, but followup works use the term \emph{saturated} to avoid confusion with asymptotic notation, and because it was previously introduced in a similar sense by \citet{karpathy2015visualizing}.} Saturation is a simplifying assumption that can be applied to a neural network that aims to exclude these degenerate representations that we might take to be impractical or unlearnable. This is similar in spirit to \autoref{sec:binarized}, where we discussed how RNNs with binarized neurons are equivalent to finite-state automata, except that that we will build a theory of low-precision networks that allows us to compare different architectures. \citet{merrill2020linguistic} define the saturated network as follows:

\begin{definition}[Saturated network] \label{def:saturation}
Let $f(x; \theta)$ be a neural network parameterized by $\theta$. We define the saturated network $\sat f(x; \theta)$ as
\begin{equation*}
    \sat f(x; \theta) = \lim_{\rho \rightarrow \infty} f(x; \rho \theta) .
\end{equation*}
\end{definition}

We can view $\sat$ as a \emph{saturation operator} mapping network functions to saturated network functions.
By taking the norm of the affine transformations in the network to infinity, saturation converts squashing functions like $\sigma$ and $\tanh$ to step functions, effectively bounding the precision of the network representations. This turns the network into a discrete automaton-like device that can be analyzed in formal-language-theoretic terms.

For example, consider a single neuron $\sigma(wx)$, for input and weight vectors $x, w \in \mathbb{Q}^m$. The saturated version of this neuron is:
\begin{equation} \label{eq:neuron}
    \lim_{\rho \rightarrow \infty} \sigma(\rho wx) .
\end{equation}
If $wx > 0$, then this limit goes to $1$. If $wx < 0$, then it goes to $0$.\footnote{When $wx = 0$, the limit is undefined. This is not a problem, since that case is measure-0. We can simply restrict the set of saturated networks to the cases where the limit is defined.} Thus, the neuron in \eqref{eq:neuron} reduces to a discrete cell whose output is either $0$ or $1$ according to a step function activation. The decision boundary is a hyperplane specified by the vector $w$. In general, saturation applies a similar analysis to \emph{every} part of the network. While simple neurons will reduce in the way described here, more complex components, such as LSTM gates, reduce to different types of discrete structures. This allows us to view the saturated versions of networks as automata, i.e., discrete machines consuming strings. Past work has explored the capabilities of these saturated networks as both language acceptors and encoders producing latent states. It has also tested how the formal properties of saturated models relates to the capabilities of unsaturated networks.

\subsection{Saturated networks as automata} \label{sec:saturated-nets}

\citet{merrill2020linguistic} conjectures that the saturated approximation of the network somehow captures the network behavior that is most stable or learnable. Empirically, this is supported by the fact that the ability of trained networks to learn formal language tasks is often predicted by the capacity of the saturated version of that network. We now go through theoretical results for various architectures, and cite relevant experimental results testing whether unsaturated networks agree with these predictions.
We will denote a saturated architecture by prefixing it with s, e.g., sLSTM.

\paragraph{RNNs} Elman sRNNs, as well as sGRUs, are equivalent to finite-state automata \citep{merrill-2019-sequential}. Experimentally, these models cannot reliably recognize $a^nb^n$ \citep{weiss2018}, Dyck languages \citep{Suzgun_2019}, or reverse strings \citep{Hao_2018, Suzgun_2019}. This agrees with the saturated capabilities, since all of these tasks require more than finite memory.

% \sat LSTMs, on the other hand, closely resemble counter automata. Experimentally, this theoretical difference is reflected by the fact that trained LSTMs can consistently learn counter languages like $a^nb^n$ \citep{Weiss2018OnTP} or $D_1$ \citep[1-Dyck; ][]{Suzgun_2019} better than RNNs or GRUs. All RNN varieties cannot learn to reverse strings \citep{Hao_2018, suzgun2019memoryaugmented}, a task requiring more memory than saturated RNNs have. Similarly, none of these RNN architectures can learn to model Dyck languages with $k > 1$ \citep{suzgun2019memoryaugmented}, a task that is similarly beyond the capacity of \sat LSTMs. \citet{shibata2020lstm} show that LSTM language models trained on natural language acquire semi-saturated representations where the gates tightly cluster around discrete values. Thus, \sat LSTMs appear to be a promising formal model of the behavior of LSTMs on both synthetic and natural tasks.

\paragraph{LSTMs} sLSTMs closely resemble counter automata \citep{merrill-2019-sequential}. Experimentally, this is reflected by the fact that trained LSTMs can consistently learn counter languages like $a^nb^n$ \citep{Weiss2018OnTP} or $D_1$ \citep[1-Dyck; ][]{Suzgun_2019}, unlike RNNs or GRUs. LSTMs cannot learn to model $k$-Dyck languages for $k > 1$ \citep{suzgun2019memoryaugmented}, which matches the fact that sLSTMs (and counter machines) do not have enough memory to model these languages. Similarly, they cannot reliably reverse strings \citep{grefenstette2015learning, Hao_2018}. \citet{shibata2020lstm} show that LSTM language models trained on natural language acquire semi-saturated representations where the gates tightly cluster around discrete values. Thus, sLSTMs appear to be a promising formal model of the counting behavior of LSTMs on both synthetic and natural tasks.

\paragraph{Stack RNNs} A stack-augmented sRNN \citep{suzgun2019memoryaugmented} becomes the deterministic stack automaton we studied in \autoref{sec:data-structures} \citep{merrill-2019-sequential}. Experimentally, trained stack-augmented sRNNs can reverse strings \citep{Hao_2018} and model higher-order Dyck languages \citep{suzgun2019memoryaugmented}, unlike the vanilla sRNN, sGRU, and sLSTM. This suggests that the expanded capacity of the saturated versions of these models translates to an expanded set of formal tasks that are learnable.

\paragraph{CNNs} On the other hand, convolutional networks (saturated or unsaturated) are less expressive than general finite-state acceptors \citep{merrill-2019-sequential}. For example, no sCNN can recognize the language $a^*ba^*$. On the other hand, the sCNN capacity is a superset of the strictly local languages \citep{merrill-2019-sequential}, which are of interest for modeling the phonology of natural language \citep{Heinz2011TierbasedSL}.

\paragraph{Transformers}
Saturation imposes interesting restrictions on the attention patterns in an sTransformer \citep{merrill-2019-sequential, merrill2020parameter}. Recall that self attention is parameterized by queries $Q$, keys $K$, and values $V$. The computation of the attention heads in an sTransformer reduces to:
\begin{enumerate}
    \item The head at position $i$ first selects the subsequence of keys $K^*$ that maximize the similarity $q_i \cdot k_j$.
    \item The output of the head is computed as the mean of the value vectors $V^*$ associated with $K^*$.
\end{enumerate}
If there is only one maximal key, then this reduces to hard attention. However, if there is a ``tie'' between several keys, then saturated attention can attend across unboundedly many positions. To see this, imagine a saturated transformer where $k_j$ is constant for all $j$. Then, attention is distributed uniformly over all positions, reducing the attention computation to an average over all value vectors. This enables the sTransformer to implement a restricted form of counting that can be used to recognize languages like $a^nb^n$. For $a^nb^n$, two heads can be used to count the number of $a$'s and $b$'s, and then the feedforward layer can verify that the quantities are the same. \citet{bhattamishra2020ability} show experimentally that unsaturated transformers learn to solve simple counting tasks using variants of this strategy. \citet{merrill2020parameter} also find evidence of counter heads in transformer language models. In summary, like the sLSTM, the sTransformer can count, predicting the fact that unsaturated transformers learn to count to solve synthetic tasks in practice.

\subsection{Space complexity}

One of the useful consequences of saturation is that it allow us to compare the memory available to different RNN representations in a formally precise framework.
By memory, we mean the amount of information that can be stored in the hidden state of the sRNN at any given step.
This analysis shifts the focus away from viewing saturated networks as acceptors, instead asking what kinds of \emph{representations} they can encode.

\citet{merrill-2019-sequential} introduce the notion of \emph{state complexity}: the number of configurations the saturated hidden state can take ranging over all inputs of length $n$. It is useful to write it in asymptotic ``big-$\bigo$'' notation, e.g., $\bigo(n)$. \citet{merrill2020linguistic} reframe this as \emph{space complexity}: the number of bits needed to represent the hidden state, which (still expressed in big-$\bigo$ notation) is essentially the base-2 log of the previous quantity.\footnote{There are some technical edge cases that separate the definitions in these two papers.} For consistency, we will discuss everything in terms of space complexity here. Space complexity provides a useful metric for describing the computational differences between different sRNNs.

The finite-state models like the sCNN, sRNN, and sGRU have $\bigo(1)$ space complexity. Due to a its counter memory, the sLSTM is $\bigo(\log n)$. This is enough to count, but not enough to encode representations like trees. In this sense, sLSTMs can be viewed as learned streaming algorithms: models that process a sequence in one pass with limited memory. On the other hand, saturated stack-augmented RNNs have $\bigo(n)$ space: enough to build unbounded hierarchical representations over the input sequence. For transformers, space complexity results may vary depending on assumptions about positional embeddings and other architectural properties. Crystalizing the space complexity of transformers is an interesting open question.\footnote{See \citet{merrill-2019-sequential} and \citet{merrill2020linguistic} for basic analysis of transformer space complexity.}

\begin{table}
    \centering
    \begin{tabular}{l|c|c|c|c|c}
        \textbf{Architecture} & sCNN & sRNN & sGRU & sLSTM & Stack sRNN \\
        \textbf{Complexity} & $\bigo(1)$ & $\bigo(1)$ & $\bigo(1)$ & $\bigo(\log n)$ & $\bigo(n)$
    \end{tabular}
    \caption{Space complexities of some saturated architectures, measured in bits. Results are taken from \citet{merrill-2019-sequential}.}
    \label{tab:complexities}
\end{table}

\subsection{Connection to learning}

Why should the capacity of saturated networks be meaningful at all for what is learnable by a network? One hypothesis is that optimization methods like stochastic gradient descent (SGD) have an inductive bias towards saturated networks. \citet{merrill2020parameter} argues that such an inductive bias should result from the continued norm growth that is a part of training. This idea is inspired by recent works in deep learning theory \citep[e.g.,][]{li2019exponential} exploring the divergence of network parameters during training. In line with this view, \citet{merrill2020parameter} show that the parameter norm of T5 \citep{raffel2020exploring} naturally drifts away from the origin over training at a rate $\sim \sqrt{t}$.\footnote{This is the same growth rate one would expect for a random walk.}
Following from \autoref{def:saturation}, the saturated capacity corresponds roughly to the class of networks that exist stably as the norm grows proportional to a scalar. Thus, continued norm growth during a long training process should guide networks towards saturated networks.\footnote{This intuition is complicated by trends in step size. For example, if step size decays exponentially over time, then the parameter norm found by training will converge, even if the norm is monotonically increasing.} To test this, \citet{merrill2020parameter} measures the difference in saturation between randomly initialized transformers and pretrained transformers. 
Whereas the randomly initialized transformer exhibit $0$ cosine similarity to saturated transformers, the representations inside the pretrained transformers are highly similar (though the similarity is $< 1$).
\citet{merrill2020parameter} also find that smaller transformer language models converge to a saturation level of $1.00$ early in training, meaning that their attention patterns essentially implement the specific variant of hard attention that exists in saturated transformers.
\section{Rational recurrences}

An alternative automata-theoretic perspective on RNNs is the framework of rational recurrences \citep{peng-etal-2018-rational, schwartz-etal-2018-bridging}. Rather than analyzing the language expressiveness and space complexity of RNN architectures, \citet{peng-etal-2018-rational} focus on the gating of the memory mechanism, i.e., the types of functions that can be computed by the hidden unit $h_t$ as a function of $x_{:t}$ and $h_{t-1}$. In particular, they define \emph{rational recurrences} as RNN gating functions $h_t$ where each element $[h_t]_i$ can be simulated by a WFA over $x_{:t}$.\footnote{Recall that we discussed theory about WFAs in \autoref{sec:wfas}.}
Crucially, they show that QRNNs, CNNs, and several other RNN variants are rational recurrences. Aside from being illuminating on a theoretical level, one practical implication of the rational recurrences framework is for efficiency, as the restricted form of a WFA makes the recurrent part of the computation potentially cheaper than for an arbitrary RNN.

One simple example of a rational recurrence is the unigram RNN. Interestingly, the equations for the unigram RNN can be ``compiled'' from a unigram WFA similar to the one shown in \autoref{fig:sigma-CM}. Transitions in the WFA correspond to parameterized gates in the WFA. This illustrates the intimate connection between WFAs and rational recurrences, and how the rational recurrences framework can motivate developing new RNNs from automata models. As an exercise to the reader: say we want to generalize the unigram RNN/WFA to condition its updates on bigrams. How would the WFA graph and RNN equations change?

While \citet{peng-etal-2018-rational} conjectured that more complicated RNNs like LSTMs and GRUs were not rationally recurrent, this question was left open in their original papers. However, \citet{merrill-etal-2020-formal} proved sLSTMs are not rationally recurrent by constructing a function computable by the sLSTM whose Hankel matrix has infinite rank. In contrast, sRNNs and sGRUs are rationally recurrent, following from their equivalence to finite-state machines. However, they need not be efficiently rationally recurrent, in the sense that the number of states required in the WFA simulating them can be very large.
\section{Transformers}

Starting with \citet{vaswani2017attention}, NLP as a field has gradually moved towards using transformers instead of LSTMs.
While attention mechanisms were originally viewed as interpretable, the highly distributed nature of transformer networks makes their inner workings mysterious.
Thus, formal analysis of transformer networks is a very interesting question that could shed light on their capacity, inductive biases, or suggest principled methods of interpretability.

We have already discussed saturated transformers in \autoref{sec:saturated-nets}. Here, we will mention analysis of several other transformer variants. Hopefully the incomplete discussion in this section will serve as an inspiration for future investigations.

From a formal perspective, a key component of the transformer architecture is its positional encodings. Without positional encodings, the transformer is barely sensitive to position at all. Even simple regular languages like $a^*b$ would be impossible to recognize, since the transformer would have no way to detect relative positional dependencies \citep{merrill-2019-sequential}.

Whereas saturated transformers can spread attention equally over a subsequence of positions \citep{merrill2020parameter}, 
\citet{hahn2020theoretical} explores the limitations of transformer networks where attention can only target a bounded number of positions. Adapting arguments from circuit complexity theory, \citet{hahn2020theoretical} shows that these hard attention transformers cannot recognize parity or Dyck languages. A similar probabilistic result holds for the soft transformer.
% This hard attention analysis differs from saturated transformers in its assumptions. Rather than allowing for ``tied'' attention over unbounded subsequences, \citet{hahn2020theoretical} explicitly restricts the support of the attention function to a finite subsequence. This explains why the conclusions conflict with those presented for the s-Transformer.

\subsection{Random feature attention}

One of the challenges with transformers is the scalability of attention for large sequence lengths. Say we have $N$ queries, each attending over a sequence of $M$ keys and values. Then the time complexity of computing attention for a fixed query is $\mathcal O(M)$, and $\mathcal O(MN)$ across all queries. In contrast, \citet{peng2021random} develop random feature attention (RFA): a drop-in replacement for attention that can be computed in time $\mathcal O(M + N)$. The name comes the fact that they use a \emph{random feature} approximation, which is derived from the unbiased estimator in \eqref{eq:estimate}:
\begin{theorem}[\citealt{NIPS2007_013a006f}]
For $1 \leq i \leq D$, let $w_i \sim \mathcal N(0, \sigma^2 I_d)$, let $\phi : \mathbb{R}^d \to \mathbb{R}^{2D}$ be a (random) nonlinear transformation of $x \in \mathbb{R}^d$:
\begin{equation}
    \phi(x) = \sqrt{1/D} \big[ \sin(w_1 \cdot x), \cdots, \sin(w_D \cdot x), \cos(w_1 \cdot x), \cdots, \cos(w_D \cdot x) \big ] .
\end{equation}
Then, the following holds:
\begin{equation}
    \mathbb{E}_{w_i} \big[ \phi(x) \cdot \phi(y) \big] = \exp \left( -\frac{\norm{x - y}^2}{2\sigma^2} \right) . \label{eq:estimate}
\end{equation}
\end{theorem}
At a high level, softmax is computed in terms of various exponential quantities resembling the right-hand side in \eqref{eq:estimate}. \citet{peng2021random} derive RFA by applying this approximation to simplify the attention computation graph.

\section{Open questions}

Finally, I will leave a non-exhaustive list of interesting guiding questions for ongoing formal methods research in modern NLP. There are a lot of potential questions here---these are things that I at least have found interesting and have maybe worked on a bit. In many cases, they relate to important issues in core NLP, such as interpretability and dataset artifacts.

\begin{enumerate}
    \item As we saw, there are many unknowns about the transformer architecture. Can we get some more complete theory about its representational capacity or inductive bias?
    \item Can modern deep learning motivate alternative theories of linguistic complexity to the Chomsky hierarchy? In particular, can we build a non-trivial theory of complexity not just for large $n$, but for bounded-length strings?
    \item How can we leverage formal models of neural network architectures for more interpretable machine learning?
    \item What are the theoretical limits on learning latent structure (syntactic or semantic) from sets of strings?
    \item Many NLP researchers believe that a major obstacle for building and evaluating systems is that models sometimes exploit spurious correlations and artifacts in datasets to generalize in ``shallow'' ways. Can we use formal language theory to distinguish spurious linguistic patterns from valid ones?
    % \item In deep learning theory, neural tangent kernels \citep[NTKs,][]{jacot2020neural} have been extensively studied to understand the behavior and generalization of infinite-width networks. Can NTKs of NLP models reveal anything about their abilities as formal language recognizers?
\end{enumerate}

I hope you have found this document interesting! Please reach out if you have thoughts, feedback, or questions.

\section*{Acknowledgments}

This document heavily benefited from discussions with Gail Weiss, Michael Hahn, Kyle Richardson, and my past and present advisors: Dana Angluin, Bob Frank, Yoav Goldberg, Noah Smith, and Roy Schwartz. Further thanks to members of Computational Linguistics at Yale, researchers at the Allen Institute for AI, and attendees and organizers of the Deep Learning and Formal Languages workshop at ACL 2019, as well as the Formalism in NLP meetup at ACL 2020.

% TODO: Acknowledge Kyle R, and anyone else who took a look at this.

\bibliography{main.bib}

\end{document}